%% file: main.tex
\documentclass[10pt,twocolumn,letterpaper]{article}

\usepackage{iccv}
\usepackage{times}
\usepackage{epsfig}
\usepackage{graphicx}
\usepackage{amsmath}
\usepackage{amssymb}
\usepackage{authblk}

\usepackage[pagebackref=true,breaklinks=true,letterpaper=true,colorlinks,bookmarks=false]{hyperref}

\iccvfinalcopy 

\def\iccvPaperID{3120} 

\ificcvfinal\fi
\begin{document}

\title{Compositional Video Prediction}
\makeatletter
\renewcommand\AB@affilsepx{ \qquad \protect\Affilfont}
\newcommand{\printfnsymbol}[1]{%
  \textsuperscript{\@fnsymbol{#1}}%
}
\makeatother



\author{Yufei Ye\textsuperscript{1}  \qquad Maneesh Singh\textsuperscript{2} \qquad Abhinav Gupta\textsuperscript{13*} \qquad Shubham Tulsiani\textsuperscript{3*} \\
\vspace{-2mm}
\textsuperscript{1}Carnegie Mellon University  \qquad \textsuperscript{2}Verisk Analytics \qquad \textsuperscript{3}Facebook AI Research \\
\vspace{1mm}
{\tt \small \{yufeiy2,abhinavg\}@cs.cmu.edu} \qquad \tt \small maneesh.singh@verisk.com  \qquad  \tt \small shubtuls@fb.com
\\
{\tt \small \href{https://judyye.github.io/CVP/}{https://judyye.github.io/CVP/}}
}

\maketitle

\begin{abstract}
We present an approach for pixel-level future prediction given an input image of a scene. We observe that a scene is comprised of distinct entities that undergo motion and
present an approach that operationalizes this insight. We implicitly predict future states of independent entities while reasoning about their interactions, and compose future video frames using these predicted states. We overcome the inherent multi-modality of the task using a global trajectory-level latent random variable, and show that this allows us to sample diverse and plausible futures. We empirically validate our approach against alternate representations and ways of incorporating  multi-modality. We examine two datasets, one comprising of stacked objects that may fall, and the other containing videos of humans performing activities in a gym, and show that our approach allows realistic stochastic video prediction across these diverse settings. See \href{https://judyye.github.io/CVP/}{project website} for video predictions.

\end{abstract}
\let\thefootnote\relax\footnotetext{*  The last two authors were equally uninvolved.}

\section{Introduction}
\input{intro}

\section{Related Work}
\input{related_work}

\section{Approach}
\input{approach}

\section{Experiments}
\input{experiment}

\section{Discussion}
\input{discussion.tex}

\noindent \textbf{Acknowledgements.} 
This research is partly sponsored by ONR MURI N000141612007 and the ARO W911NF-18-1-0019.

{\small
\bibliographystyle{ieee_fullname}
\bibliography{ref}
}

\clearpage
\appendix
\input{appendix.tex}
\end{document}

%% file: intro.tex

A single image of a scene allows us humans to make a remarkable number of judgments about the underlying world. For example, consider the two images on the left in Fig \ref{fig:teaser}. We can easily infer that the top image depicts some stacked blocks, and the bottom shows a human with his arms raised. While these inferences showcase our ability to understand what is, even more remarkably, we are capable of predicting what will happen next. For example, not only do we know that there are stacked blocks in the top image, we understand that the blue and yellow ones will topple and fall to the left. Similarly, we know that the person in the bottom image will lift his torso while keeping his hands in place. In this work, we aim to build a model that can do the same -- from a \emph{single} (annotated) image of a scene, predict at a pixel level, what the future will be.

\input{teaser.tex}

A key factor in the ability to make these predictions is that we understand scenes in terms of `entities', that can move and interact e.g.\ the blocks are separate objects that move; the human body's motion can similarly be understood in terms of the correlated motion of the limbs. We operationalize this ideology and present an approach that instead of directly predicting future frames, learns to predict the future locations and appearance of the entities in the scene, and via these composes a prediction of the future frame. The modeling of appearance and the learned composition allows our method to leverage the benefits of independent per-entity representations while allowing for reasoning in pose changes or overlap/occlusions in pixel space.


Although our proposed factorization allows learning models capable of predicting the future frames via entity-based reasoning, this task of inferring future frames from a single input image is fundamentally ill-posed.  To allow for the inherent multi-modality of the prediction space, we propose to use a trajectory-level latent random variable that implicitly captures the ambiguities over the whole video and train a future predictor conditioned of this latent variable. We demonstrate that modeling the ambiguities using this single latent variable instead of per-timestep random variables allows us to make more realistic predictions as well as sample diverse plausible futures.

We validate our approach using two datasets where the `entities' either represent distinct objects, or human body joints, and demonstrate that the same method allows for predicting future frames across these diverse settings. We demonstrate: (a) the benefits of our proposed entity-level factorization; (b) ability of the corresponding learned decoder to generate future frames; (c) capability to sample different futures.

%% file: teaser.tex
\begin{figure}[t]
    \centering
    \includegraphics[width=.48\textwidth]{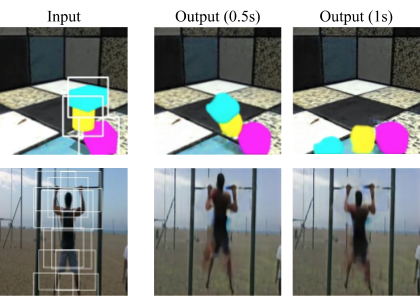}
    \caption{Given a still image with locations of entities (objects or joints), we predict a sequence of future frames. We visualize two frames from the predicted sequence for the given inputs.}
    \label{fig:teaser}
\end{figure}

%% file: related_work.tex
\paragraph{Modeling Physical Interaction.}
Many recent works ~\cite{watters2017visual, kipf2018neural, battaglia2016interaction, santoro2017simple, chang2016compositional, hsieh2018learning} study modeling multiple objects in physical systems. Similar to us, they reason using the relationship between objects, and can predict the trajectories over a long time horizon. However, these approaches typically model deterministic processes under simple visual (or often only state based) input, while often relying on observed sequences instead of a single frame. Although some recent works take raw image as input \cite{watters2017visual, fragkiadaki2015learning}, they also only make prediction in state, and not pixel space.
In contrast to these approaches, while we also use insights based on modeling physical interaction, we show results for video frame generation in a stochastic setup, and therefore also need to (implicitly) reason about other properties such as shape, lighting, color. Lastly, a related line of work is to predict stability of configurations~\cite{li2016visual, groth2018shapestacks, li2016fall, jia20153d, lerer2016learning}. Our video forecasting task also requires this understanding, but we do not pursue this as the end goal.
\vspace{-4mm}
\paragraph{Video Factorization.}
It is challenging to directly predict pixels due to high dimensionality of the prediction space, and several methods have been used to factorize this output space~\cite{vondrick2016generating, villegas2017decomposing, tulyakov2017mocogan, denton2017unsupervised}. The main idea is to separate dynamic foreground from static background and generate pixels correspondingly. While these approaches show promising results to efficiently model one object motion, we show the benefits of modeling multiple entities and their interactions.

Another insight has been to instead model the output space differently, e.g.\ optical flow \cite{walker2016uncertain, liu2017video}, or motion transformation~\cite{xue2016visual, chen2017video, jia2016dynamic, finn2016unsupervised}. This enables generating more photo-realistic images for shorter sequences, but may not be applicable for longer generation as new content becomes visible, and we therefore pursue direct pixel generation. Another line of work proposes to predict future in a pre-defined structured representation space, such as human pose ~\cite{walker2017pose,villegas2017learning}. While our approach also benefits from predicting an intermediate structured representations, it is not our end goal as we aim to generate pixels from this representation.
\input{pipeline.tex}

\vspace{-4mm}
\paragraph{Object-centric video prediction.}
A line of work explicitly enumerates the state of each object as location, velocity, mass, etc, then applies planning algorithm to unroll movement under reward ~\cite{kitani2012activity,huang2015approximate}, or leverage Newtonian dynamics~\cite{ye2018interpretable, wu2016physics}. However, these explicit representation based methods may not be applicable when the state space is hard to define, or pixel-wise predictions are not easily inferred given such a state e.g.\ human motions on complex background.

\vspace{-4mm}
\paragraph{Stochastic prediction.}
Predicting the future is an inherently multi-modal task. Given a still image or a sequence of frames, there are multiple plausible futures that could happen. The uncertainty is usually encoded as a sequence of latent variables, which are then used in a generative model such as GAN~\cite{goodfellow2014generative} based~\cite{mathieu2015deep, vondrick2016generating, chen2017video, tulyakov2017mocogan}, or, similar to ours, VAE~\cite{kingma2013auto} based~\cite{walker2016uncertain, denton2018stochastic}. These methods~\cite{fragkiadaki2017motion,denton2018stochastic,yanMTVAE} often leverage an input sequence instead of a single frame, which helps reduce the ambiguities. Further, the latent variables are either per-timestep~\cite{denton2018stochastic}, or global~\cite{babaeizadeh2017stochastic,yanMTVAE} whereas our model leverages a global latent variable, which in turn induces per-timestep variables.


%% file: pipeline.tex
\begin{figure*}[t]
    \centering
    \includegraphics[width=\textwidth]{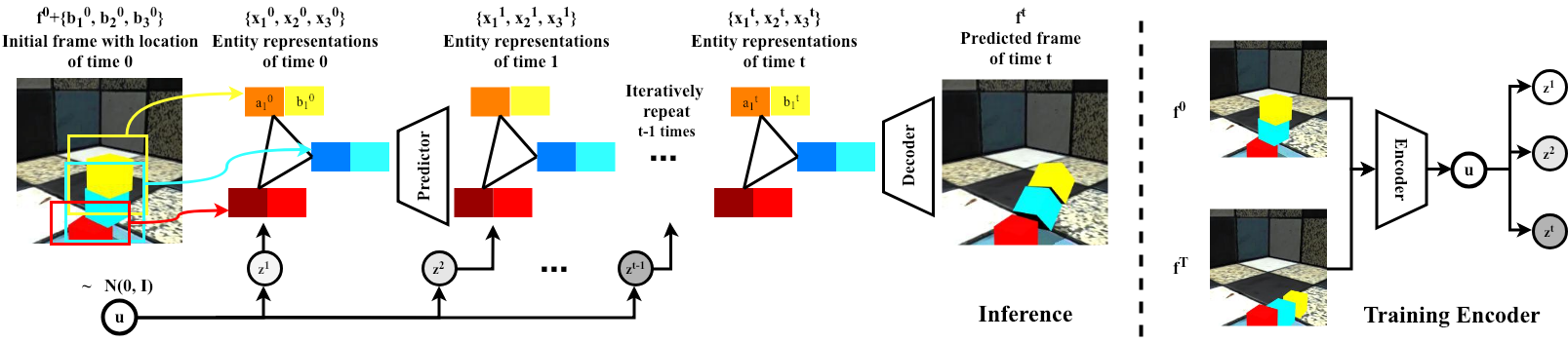}
    \caption{Our model takes as input an image with known/detected location of entities. Each entity is represented as its location and an implicit feature. Given the current entity representations and a sampled latent variable, our prediction module predicts the representations at the next time step. Our learned decoder composes the predicted representations to an image representing the predicted future. During training, a latent encoder module is used to infer the distribution over the latent variables using the initial and final frames.}
    \label{fig:pipeline}
\end{figure*}


%% file: approach.tex
Given an input image along with (known or detected) locations of the entities present, our goal is to predict a sequence of future frames. Formally, given a starting frame $f^0$ and the location of $N$ entities $\{b^0_n\}_{n=1}^N$, we aim to generate $T$ future frames $f^1, f^2, ..., f^T$.
This task is challenging mainly for two reasons: a) the scene may comprise of multiple entities, making it necessary to account for their different dynamics and interactions, and b) the inherently multi-modal nature of the prediction task.

To overcome the first challenge, our insight is that instead of modeling how the scene changes as a whole, we should pursue prediction by modeling how the entities present change.  We do so using an \emph{entity predictor} that predicts per-entity representations: $\{x_n^t\}_{n=1}^N \equiv \{(b_n^t, a_n^t)\}_{n=1}^N$, where $b_n^t$ denotes the predicted location, and $a_n^t$ denotes predicted features that implicitly capture appearance for each entity. While this factorization allows us to efficiently predict the future in terms of these entities, an additional step is required to infer pixels. We do so using a \emph{frame decoder} that is able to retain the properties of each entity, respect the predicted location, while also resolving the conflicts e.g.\ occlusions when composing the image.

To account for the fundamental multi-modality in the task, we incorporate a global random latent variable $u$ that implicitly captures the ambiguities across the whole video. This latent variable $u$, in turn deterministically (via a learned network) yields per-timestep latent variables $z_t$ which aid the per-timestep future predictions. Concretely, the predictor $\mathcal{P}$ takes as input the per-entity representation $\{x_n^t\}$ along with the latent variable $z_t$, and predicts the entity representations at the next timestep $\{x_n^{t+1}\} \equiv \mathcal{P}(\{x_n^t\}, z_t)$. The decoder $\mathcal{D}$, using these predictions (and the initial frame $f^0$ to allow modeling background), composes the predicted frame $f^t \equiv \mathcal{D}(\{x_n^t\}, f^0)$.

We train our model to maximize the likelihood of the training sequences, comprising of terms for both the frames and the entity locations. As is often the case with optimizing likelihood in models with unobserved latent variable models e.g.\ VAEs~\cite{kingma2013auto}, directly maximizing likelihood is intractable, and we therefore maximize a variational lower bound. Towards this, we train another module, a \emph{latent encoder}, which predicts a distribution over the latent variable $u$ using the target video. Note that the annotation of future frames/locations, as well as the latent encoder, are all only used during training. During inference, however, as illustrated in Fig \ref{fig:pipeline}, we take in input only a single frame along with (predicted/known) locations of the entities present, and can generate multiple plausible future frames. We first describe the predictor, decoder, and encoder modules in more detail, and the present the overall training objective.

\input{predictor}

\input{decoder}

\input{encoder}



%% file: predictor.tex
\subsection{Entity Predictor}
\label{sec:appr:predictor}
Given per-entity locations and implicit appearance features $\{x_n^t\}_{n=1}^N \equiv \{(b_n^t, a_n^t)\}_{n=1}^N$, the predictor outputs the predictions for the next time step using the latent variable $z_t$. An iterative application of this predictor therefore allows us to predict the future frames for the entire sequence using the encodings from the initial frame. To obtain this initial input to the predictor i.e.\ the entity encodings at the first time step $\{x_n^0\}_{n=1}^N$, we use the known/detected entity locations $\{b_n^0\}$, and extract the appearance features $\{a_n^0\}$ using a standard ResNet-18 CNN~\cite{he2016deep} on the cropped region from $f^0$.

While the predictor $\mathcal{P}$ infers per-entity features, the prediction mechanism should also allow for the interaction among these entities rather than predicting each of them independently e.g.\ a block may or may not fall depending on the other ones around it. To enable this, we leverage a model in the graph neural network family, in particular based on `Interaction Networks' which take in a graph $G=(V, E)$ with associated features for each node, and update these via iterative message passing and message aggregation. See ~\cite{battaglia2018relational} for a more detailed review. Our predictor $\mathcal{P}$ that infers $\{x_n^{t+1}\}$ from $(\{x_n^{t}\}, z_t)$ comprises of 4 interaction blocks, where the first block takes as input the entity encodings concatenated with the latent feature: $\{x_n^t \oplus z_t\}_{n=1}^N$. Each of these blocks performs a message passing iteration using the underlying graph, and the final block outputs predictions for the entity features for the next timestep $\{x_n^t\}_{n=1}^N \equiv \{(b_n^t, a_n^t)\}_{n=1}^N$. 
This graph can either be fully connected as with our synthetic data experiments, or more structured e.g.\ skeleton in our human video prediction experiments. See appendix for more details on the message passing operations.

Although our prediction module falls under the same umbrella as Interaction Networks(IN) \cite{battaglia2016interaction}, which are in turn related to Graph Convolution Networks(GCN) \cite{kipf2018neural}, there are subtle differences, both in the architecture and application. While ~\cite{battaglia2016interaction} use a single interaction block to update node features, we found that stacking multiple interaction blocks for each timestep is particularly helpful. In contrast to GCNs which use a predefined mechanism to compute edge weights and use linear operations for messages, we find that using non-linear functions as messages allows better performance. Finally, while existing approaches do apply variants of GNNs for future prediction, these are restricted to predefined state-spaces as opposed to pixels, and do not account for uncertainties using latent variables.


%% file: decoder.tex
\subsection{Frame Decoder}
\label{sec:appr:decoder}

The decoder aims to generate pixels of the frame $f^t$ from a set of predicted entity representations. While the entity representations capture the moving aspects of the scene, we also need to incorporate the static background, and additionally use the initial frame $f^0$ to do so. Our decoder $\mathcal{D}$, as depicted in Fig \ref{fig:decoder}, therefore predicts $f^t \equiv \mathcal{D}(\{x_n^t\}, f^0)$. To compose frames from this factored input representation, there are several aspects that our decoder must  consider: a) the predicted location of the entities should be respected, b) the per-entity representations may need to be fused e.g.\ when entities occlude each other, and c) different parts of background may become visible as objects move.

To account for the predicted location of the entities when generating images, we propose to decode a normalized spatial representation for each entity, and warp it to the image coordinates using the predicted 2D locations. To allow for the occlusions among entities, we  predict an additional soft mask channel for each entity, where the value of masks are supposed to capture the visibility of the entities. Lastly, we overlay the (masked) spatial features predicted via the entities onto a canvas containing features from the initial frame $f^0$, and then predict the future frame pixels using this composed feature.

\input{fig_decoder.tex}

More formally, let us denote by $\phi_{bg}$ the spatial features predicted from the frame $f^0$ (using a CNN with architecture similar to UNet), and let $\{(\bar{\phi}_n, \bar{M}_n) = g(a_n)\}_{n=1}^N$ denote the features and spatial masks decoded per-entity using an up-convolutional decoder network $g$. We first warp, using the predicted locations $b_n$, these features and masks into image coordinates at same resolution as $\phi_{bg}$. Denoting by $\mathcal{W}$ a differentiable warping function e.g.\ in Spatial Transformer Networks \cite{jaderberg2015spatial}, we can obtain the entity features and masks in the image space:
\begin{gather}
    \phi_n = \mathcal{W}(\bar{\phi}_n, b_n);~~M_n = \mathcal{W}(\bar{M}_n, b_n)
\end{gather}
Note that the warped mask and features $(\phi_n, M_n)$ for each entity are zero outside the predicted bounding box $b_n$, and the mask $M_n$ can further have variable values within this region. Using these independent background and entity features, we compose frame level spatial features $\phi$ by combining these via a weighted average. Denoting by $M_{bg}$ a constant spatial mask (with value 0.1), we obtain the composed features as:
\begin{gather}
    \phi = \frac{\phi_{bg} \odot M_{bg}~~\oplus~~\sum_n \phi_{n} \odot M_{n}}{M_{bg}~~\oplus~~\sum_n M_{n}}
\end{gather}
These composed features $\phi$ incorporate information from all entities at the appropriate spatial locations, allow for occlusions using the predicted masks, and incorporate the information from background. We then decode the pixels for the future frame from these composed features. Note that one has a choice over the spatial level where this feature composition happens e.g.\ it can happen in feature space at near the image resolution (late fusion), or even directly at pixel level (where the variables $\phi$ all represent pixels), or alternatively at a lower resolution (mid/early fusion). We find that late fusion in implicit (and not pixel) space yields most promising results, and also find that the inferred masks end up correspond to instance segmentations.

%% file: fig_decoder.tex
\begin{figure}[t!]
    \centering
    \includegraphics[width=0.85\linewidth]{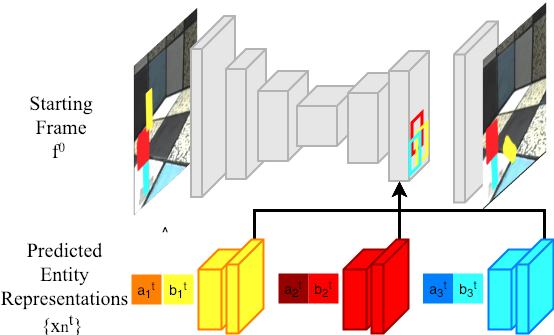}
    \caption{Our frame decoder takes in the initial frame $f^0$ and the predicted entity representations at time $t$, and outputs the frame corresponding to the predicted future $f^t$.}
    \label{fig:decoder}
    \vspace{-3mm}
\end{figure}

%% file: encoder.tex
\subsection{Latent Representation}
\label{sec:appr:encoder}
We described in Sec \ref{sec:appr:predictor} how our prediction module is conditioned on a latent variable $u$, which in turn generates per-timestep conditioning variables $z_t$ that are used in each prediction step -- this is depicted in Fig \ref{fig:encoder}(a). Intuitively, the global latent variable would capture video-level ambiguities e.g.\ where the blocks fall, the variables $z_t$ resolve the corresponding ambiguities in the per-timestep motions. While previous approaches for future prediction similarly use latent variables to resolve ambiguities (see  Fig \ref{fig:encoder}(c-d)), the typical idea is to use independent per-timestep random variables, whereas in our model the $z_t$'s are all correlated.

\input{fig_encoder.tex}

During training, instead of marginalizing the likelihood of the sequences over all possible values of the latent variable $u$, we instead minimize the variational lower bound of the log-likelihood objective. This is done via training another module, a \emph{latent encoder}, which (only during training) predicts a distribution over $u$ conditioned on the ground-truth video. In practice, we find that simply conditioning on the first and last frame of the video (using a feed-forward neural network) is sufficient, and denote by $q(u|f^0, \hat{f}^T)$ the distribution predicted. Given a particular $u$ sampled from this distribution, we recover the $\{z_t\}$ via a one-layer LSTM which, using $u$ as the cell state, predicts the per-timestep variables for the sequence.

\subsection{Training Objective}
Overall, our training objective can be viewed as maximizing the log-likelihood of the ground-truth frame sequence $\{\hat{f}^t\}_{t=1}^T$. We additionally use training-time supervision for the locations of the entities $\{\{ \hat{b}^t_n\} _{n=1}^N\}~_{t=1}^T$. While this objective has an interpretation of log-likelihood maximization, for simplicity it is  described as a loss $L$ with different terms,
where the first $L_{pred}$  encourages the frame and location predictions to match the ground-truth:
\begin{align*}
L_{pred} &= \sum_{t=1}^T (\|\mathcal{D}(\{x_n^{t}\}, f^0) - \hat f^{t}\|_1  +  \lambda_1 \sum_{n=1}^N \|b_n^t - \hat b_n^t\|^2 )
\end{align*}
The second component corresponds to enforcing an information bottleneck on the latent variable distribution:
\begin{align*}
    L_{enc} &= KL[q(u)~\|~\mathcal{N}(0, I)]
\end{align*}
Lastly, to further ensure that the decoder generates realistic composite frames, we add an auto-encoding loss that enforces it generates the correct frame when given entities representations $\{\hat{x}_n^{t}\}$ extracted from $\hat{f}^t$ (and not the ones predicted) as input.
\begin{align*}
    L_{dec} &= \sum_{t=0}^T \|\mathcal{D}(\{ \hat x_n^{t}\}, f^0) - \hat f^{t}\|_1 
\end{align*}
The total loss is therefore $L = L_{dec} +  L_{pred} + \lambda_2 L_{enc}$ with hyper-parameter $\lambda_2$ determining the trade-offs among accurate predictions and information bottleneck in random variable. See appendix for additional details. We will release our code for reproducibility.

%% file: fig_encoder.tex
\begin{figure}[t!]
    \centering
    \includegraphics[width=0.95\linewidth]{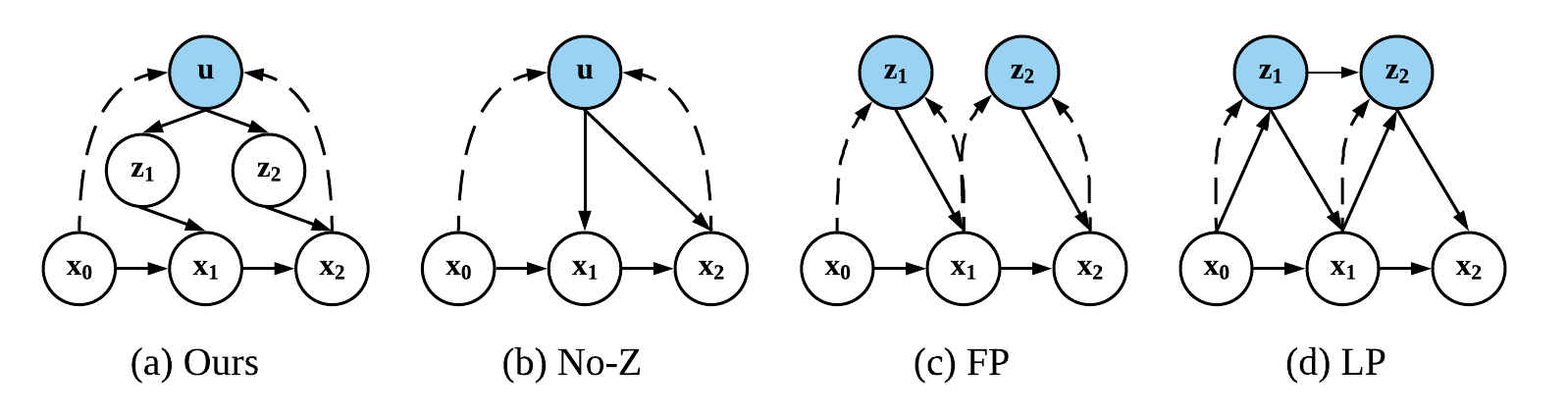}
    \caption{Our encoder (a) and baseline encoder (b-d). At test time, variables in blue are sampled randomly. At training, encoders model the posterior by all $x$s connected with dotted lines.}
    \label{fig:encoder}
    \vspace{-3mm}
\end{figure}

%% file: experiment.tex
We aim to show qualitative and quantitative results highlighting the benefits of various components (predictor, decoder, and latent representation) in our approach, and aim to highlight that our approach is general to accommodate various scenarios. See generated videos in supplementary.

\subsection{Experiment Setup}
\label{sec:sec:exp_setup}
\vspace{-1mm}
\paragraph{Dataset.}

We demonstrate our results on both the synthetic (ShapeStacks~\cite{groth2018shapestacks}) and real (Penn Action~\cite{zhang2013actemes}) dataset. 
Shapestacks is a synthetic dataset comprised of stacked objects that fall under gravity with diverse blocks and configurations. The blocks can be cubes, cylinders, or balls with different colors. 
In addition to evaluating generalization ability, we further test with similar setups with videos comprised of 4, 5 or 6 blocks.

Penn Action~\cite{zhang2013actemes} is a real video dataset of people playing various indoor and outdoor sports with annotations of human joint locations. The Penn Action dataset is challenging because of a) diverse backgrounds, view angles, human poses and scales b) noise in annotations, and c) multiple activity classes with different dynamics. 
We use a subset of the categories related to gym activities because most videos in these classes do not have camera motion and their backgrounds are similar within these categories. 
We adopt the recommended train/test split in ~\cite{zhang2013actemes}.
Beyond that, we argue it is not impractical to assume known locations -- we substitute ground truth annotation $\hat b^t_n$ with key-points location from off-the-shelf detector ~\cite{fang2017rmpe} in both training and testing. 

In both scenarios, we train our model to generate video sequence of 1 second given an initial frame, using exactly the same architecture despite the two diverse scenarios -- entities correspond to objects in Shapestacks and correspond to  joints of human body  in Penn Action. 



\vspace{-4mm}
\paragraph{Evaluation Metrics.} 
In both of these settings, we evaluate the predicted entity locations using average mean square error and the quality of generated frames using the Learned Perceptual Image Patch Similarity (LPIPS) \cite{zhang2018perceptual} metric. A subtle detail in the evaluation is that at inference, the prediction is dependent on a random variable $u$, and while only a single ground-truth is observed, multiple predictions are possible. To account for this, we draw 100 samples of latents and record the best scores as in \cite{denton2018stochastic}. When we ablate non-stochastic modules (e.g.\ decoders), we use the mean $u$ predicted by the latent encoder (after seeing the `ground-truth' video). Without further specification, the curves are plotted in the `best of 100' setup; the qualitative results visualize the best predictions in terms of LPIPS. 

\input{pred_quan.tex}
\input{pred_vis.tex}

\vspace{-4mm}
\paragraph{Baselines.}
There are three key components in our model, i.e.\ the entity predictor, frame decoder, and latent representation. Various baselines are provided to highlight our choices in each of the components. Among them, some variant specifically points to previous approaches as the following: \begin{itemize}
    \vspace{-1.5mm}\item No-Factor ~\cite{lerer2016learning} only predicts on the level of frames. Here we provide supervision from entity locations and pixels instead of segmentation masks;
    \vspace{-2.5mm}\item LP ~\cite{denton2018stochastic}  implements the stochastic encoder module in SVG-LP to compare  different dependency of latent variables; 
    \vspace{-2.5mm}\item Pose Knows \cite{walker2017pose}  is most related to our Penn Action setting which also predicts poses as intermediate representation, but it predicts location jointly and generates videos in a different way. \vspace{-1.5mm}
\end{itemize}
Besides the above which are strongly connected to previous works, we also present other baselines whose details are discussed in Section \ref{sec:sec:shapestack}.
\input{pred_num.tex}

\subsection{Analysis using Shapestacks}
\label{sec:sec:shapestack}


We use Shapestacks to validate the different components of the proposed approach i.e.\ the entity predictor, frame decoder, and the modeling choices for the latent variables. 


\vspace{-4mm}
\paragraph{Entity Predictor.}
We aim to show that our proposed predictor, which is capable of factorizing prediction over per-entity locations and appearance, as well as allowing reasoning via GNNs, improves prediction. 
Towards this, we compare against three alternate models: a) No-Factor ~\cite{lerer2016learning}, b) No-Edge and c) Nearest Neighbor(NN). The No-Factor model does not predict a per-entity appearance but simply outputs a global feature that is decoded to foreground appearance and mask.  To leverage the same supervision as box locations, it also takes as input (and outputs) the per-entity bounding boxes. The No-Edge does not allow for interactions among entities when predicting the future. The NN computes the features of the initial frame using a CNN. During inference, it retrieves the training video that is most similar in terms of those features. See appendix for details. 


Figure \ref{fig:pred_vis} shows the prediction using our model and the baselines. The No-Factor generates plausible frames at the beginning and performs well for static entities. However, at later time steps, entities with large range of motion diffuse because of the uncertainty.  In contrast, entities generated by ours have clearer boundary over time. The No-Edge does not accurately predict block orientations as it requires more information about relative configuration, and further changes the colors over time. In contrast, blocks generated by our approach gradually rotate and fall over and  are learned to retain colors. The NN baseline shows our model does not simply memorize the training set.  Figure \ref{fig:pred_quan} reports quantitative evaluations, and similarly observe the benefits of our approach.

Figure \ref{fig:pred_num} shows the results when the model generalizes to  different number of entities (4, 5, and 6) at test time.  The No-Factor uses fully connected layers to predict which cannot be directly adapted to variable number of blocks. We show methods that are able to accommodate the number of entities changes, i.e. \ No-Edge and ours. 
Our method predicts locations closer to the truth with more realistic appearance, and is able to retain the blocks color across time.  Note that we train all models with only three blocks.

\vspace{-4mm}
\paragraph{Primitive Decoder.}
\input{dec_vis.tex}
\input{dec_mask.tex}

While the No-Factor baseline shows the benefits of composing features for each entity while accounting for their predicted spatial location, we ablate here whether this composition should directly be at a pixel-level or at some implicit feature level (early, mid, or late). Across all these ablations, the number of layers in the decoder remain the same; only the level at which features from entities are composed differs.

The qualitative result are shown in Fig \ref{fig:dec_vis} where the first row visualizes decodings from the initial frame, and the second row demonstrates decoding from predicted features for a later timestep. While both late/pixel-level fusion reconstructs the initial frame faithfully, the pixel-level fusion introduces artifacts for future frames. The mid/early fusion alternates do not capture details well. We also observe similar trends in  the quantitative results visualized in  Figure \ref{fig:dec_mask}. Note the latent $u$ is encoded by the ground truth videos.


To further analyze the decoder, we visualize the generated soft masks in Figure \ref{fig:dec_mask}. The values indicate the probability of the pixel belongs to a foreground of the entity. Note that this segmentation emerges despite of no direct supervision, but only location and frame-level pixels.

\input{encBest_quan.tex}
\input{enc_vis.tex}

\vspace{-4mm}
\paragraph{Latent Representation.}
Our choice of the latent variables in the prediction model differs from the common choice of using a per-timestep random variable $z_t$. We compare our approach (Figure \ref{fig:encoder}a) with such other alternatives (Figure \ref{fig:encoder} b-e). The No-Z baseline  (Figure \ref{fig:encoder}b)  directly uses $u$ across every time step, instead of predicting a per-timestep  $z^t$ from it. In both Fixed Prior (FP) and Learned Prior(LP)~\cite{denton2018stochastic} baselines, the random variables are sampled per time step, either independently (FP), or depending on previous prediction (LP). During training, both FP and LP models are trained using an encoder similar to ours, but this encoder that predicts $z_t$ using the frames $f^t$ and $f^{t+1}$ (instead of our approach using $f^0$ and $f^{T}$ to predict $u$).
\input{gym_vis.tex}
\input{gym_sample.tex}
\input{gym_quan.tex}

We visualize using five \emph{random} samples in form of trajectories of entity locations in Figure \ref{fig:enc_vis}.
We notice that the direction of trajectories from No-Z model do not change across samples. The FP model has issues maintaining consistent motions across time-steps as during each timestep, an independent latent variable is sampled. The LP method performs well compared to FP, but still has similar issues. Compared to baselines, the use of a global latent variable allows us to sample and produce consistent motions across a video sequence, while also allowing diverse predictions across samples.
The quantitative evaluations in Figure \ref{fig:encBest_quan} show similar benefits where our method does well for both location error and frame perceptual distance over time.


\subsection{Penn Action}

Our model used in this dataset is exactly the same as that in the Shapestacks. Only the graph in the predictor is based on the human skeleton, instead of fully-connected. 
See project page for generated videos.

We compare with Pose-Knows \cite{walker2017pose} which leverages entities as intermediate representation and generates pixel-level prediction. However, they a) do not predict feature for appearance but only location of each entity (joint); b) do not involve interaction mechanism; c) adopt a different generation  method (GAN) where they stick sequence of rendered pose figures to the initial frame, and fuse them by a spatial-temporal 3D convolution network \cite{tran2015learning}. In their paper,  the adversarial loss is posed to improve realism. We present that our method also benefits from the adversarial loss (Ours+Adv).


Figure \ref{fig:gym_vis} and Figure \ref{fig:gym_quan} show qualitative and quantitative results using the best latent variable among 100 samples.
The No-Factor cannot generate plausible foreground while the No-Edge does not compose well. Our results improve to be sharper if  adversarial loss is added  (Ours+Adv). 

We also visualize predictions when, during both \textit{training} and inference, annotated key-points are replaced with detected key-points using \cite{fang2017rmpe}. We note that the performance is competitive to the setting using annotated key-point locations, indicating that our method is robust to annotation noise. It also indicates that the requirement of entity locations is not a bottle-neck, since automatically inferred location suffice in our experiment.

Figure \ref{fig:gym_sample} visualizes different sample futures using the predicted joint locations across time. Our model learns  the boundary of the human body against the background as well as  how the entities compose the human body even when they heavily overlap. More interestingly, the model learns different types of dynamics for different sports. For example, in pull ups, the legs move more while the hands are still; in clean and jerk, the legs almost remain at the same place. 


%% file: pred_quan.tex
\begin{figure}[t!]
    \centering
        \includegraphics[width=\linewidth]{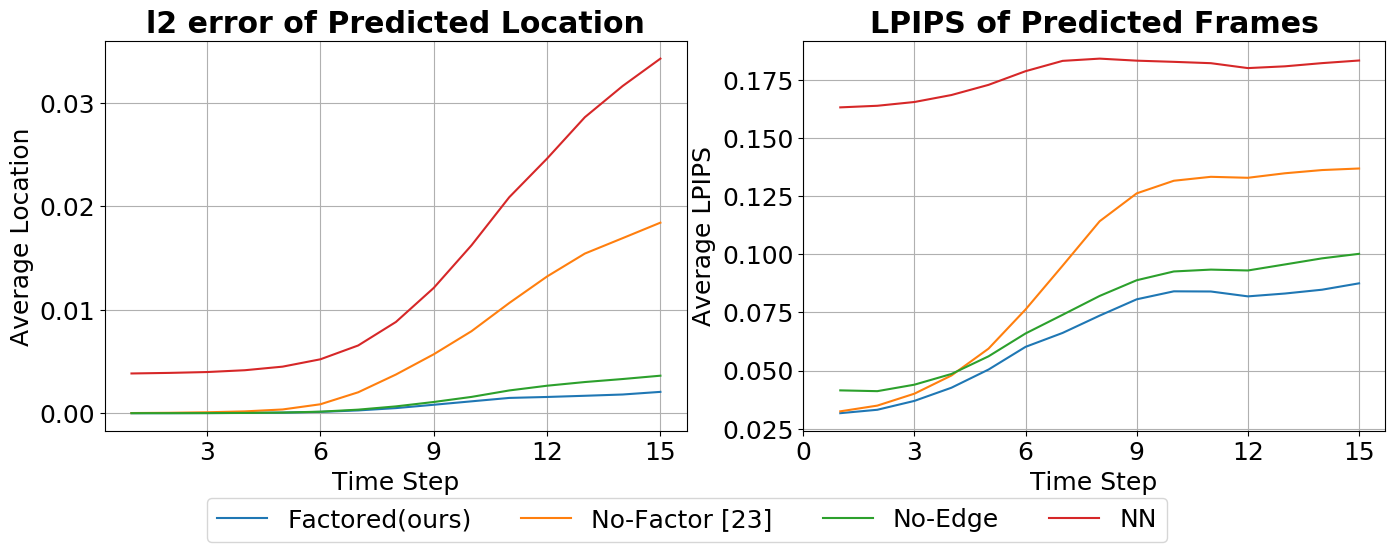}
    \caption{Error for location (Left) and frame  (Right) prediction using our entity predictor and baselines. For each sequence, the best score of 100 random samples is drawn.
    }\label{fig:pred_quan}
    \vspace{-3mm}
\end{figure}

%% file: pred_vis.tex
\begin{figure*}[t!]
        \includegraphics[width=\textwidth]{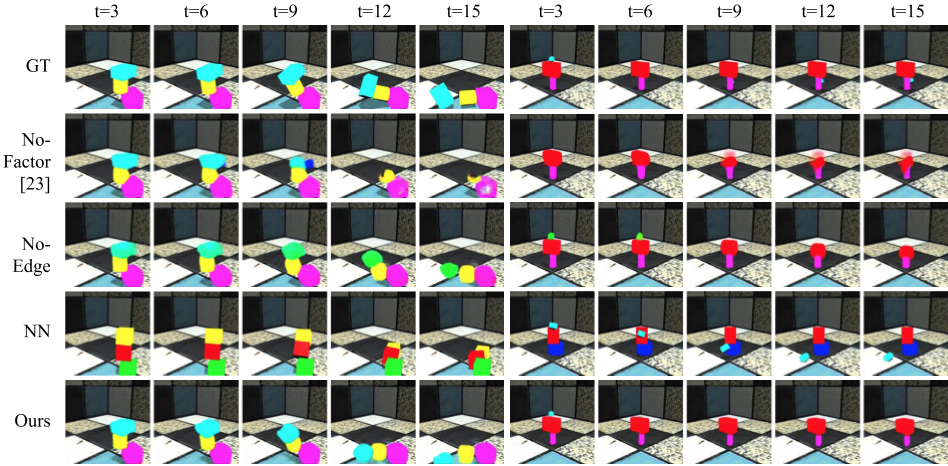}
    \caption{Video predictions using our predictor compared to baselines. We visualize the generated sequence after every 3 time steps.}
    \label{fig:pred_vis}
    \vspace{-5mm}
\end{figure*}

%% file: pred_num.tex
\begin{figure}[th]
    \centering
    \includegraphics[width=\linewidth]{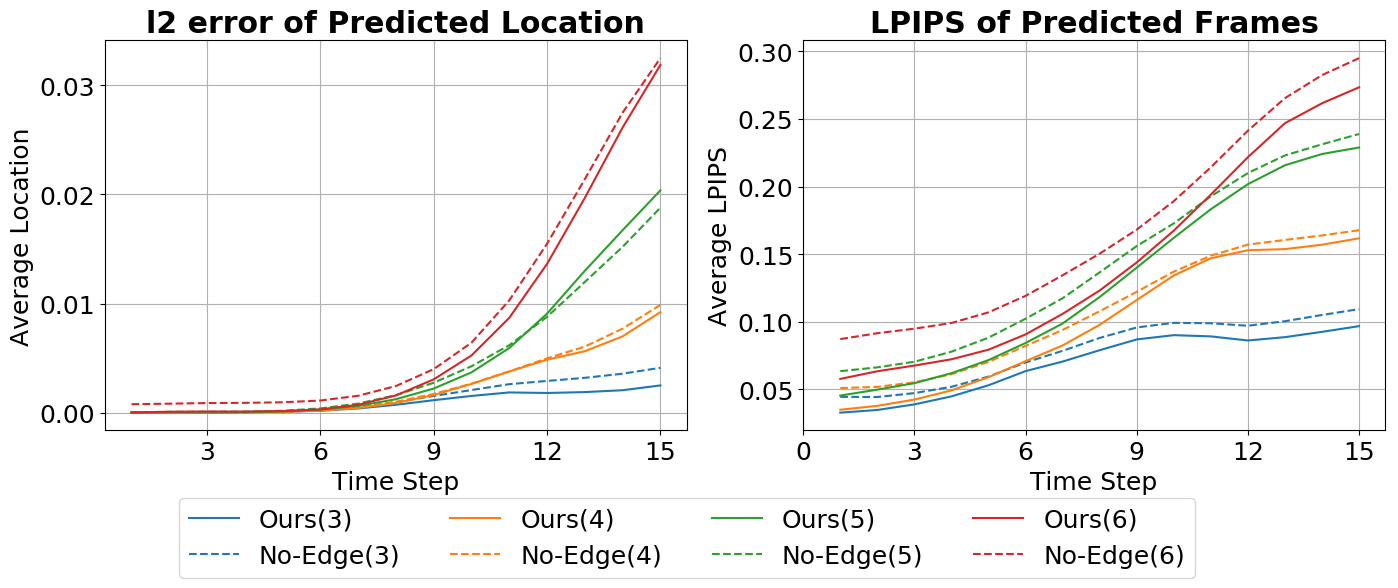}
    \includegraphics[width=\linewidth]{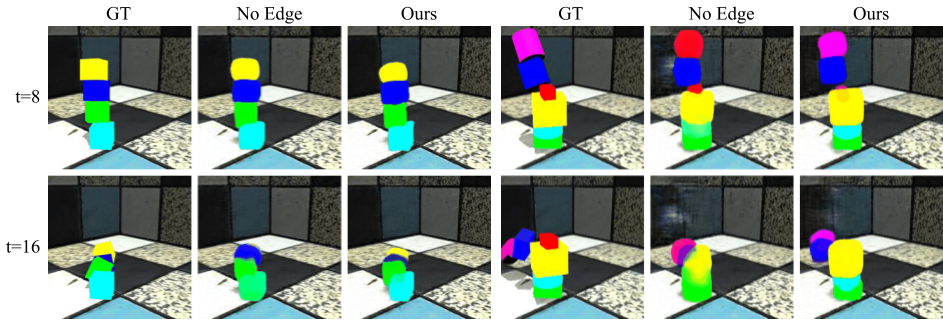}
    \caption{Above: Quantitative evaluation of the entity predictor when generalized to different number of blocks. The number in the bracket indicates the number of blocks in the subset. Below: Video predictions. The middle and last step are visualized.}
    \label{fig:pred_num}
    \vspace{-3mm}
\end{figure}

%% file: dec_vis.tex
\begin{figure}[t!]
    \centering
        \includegraphics[width=\linewidth]{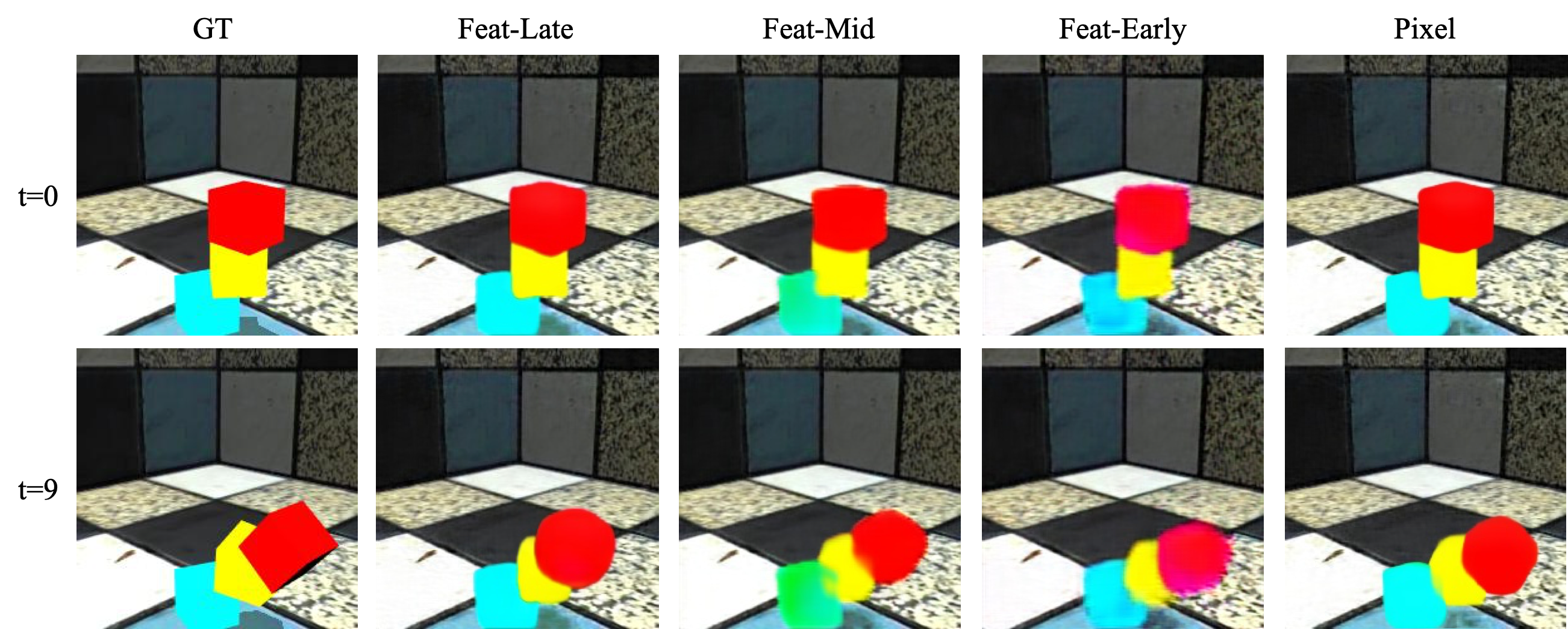}
        \caption{Qualitative results for composing entity representations into a frame. We visualize the outputs from variants of the decoder performing Late/Mid/Early fusion in feature space, or directly in pixel space. The first row depicts decoding of the initial representation; the second row depicts decoding of the predicted entities at a later time step.}
        \label{fig:dec_vis}
\end{figure}

%% file: dec_mask.tex
\begin{figure}[t!]
    \centering
        \includegraphics[width=\linewidth]{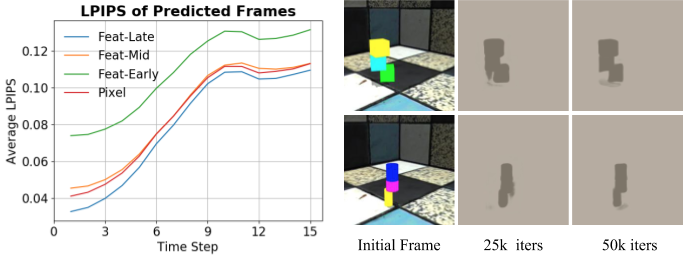}
    \caption{Left: Average Perceptual error for predicted frames via variants of the decoder. Right: Visualization of the composition of the foreground masks predicted for the entities. }
    \vspace{-3mm}
    \label{fig:dec_mask}
\end{figure}

%% file: encBest_quan.tex
\begin{figure}
    \centering
        \includegraphics[width=\linewidth]{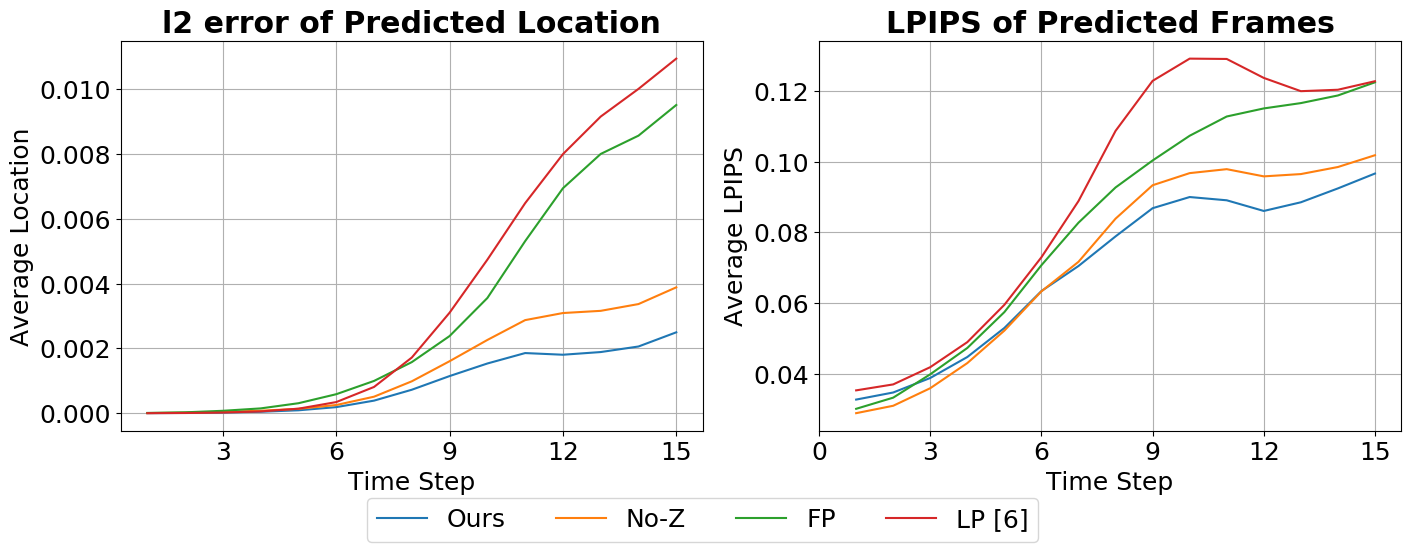}
    \caption{Error for location (Left) and frame (Right) prediction  using our encoder and baselines. For each sequence, the best score of 100 random samples is drawn.}
    \vspace{-3mm}
    \label{fig:encBest_quan}
\end{figure}

%% file: enc_vis.tex
\begin{figure}[t!]
    \centering
    \includegraphics[width=\linewidth]{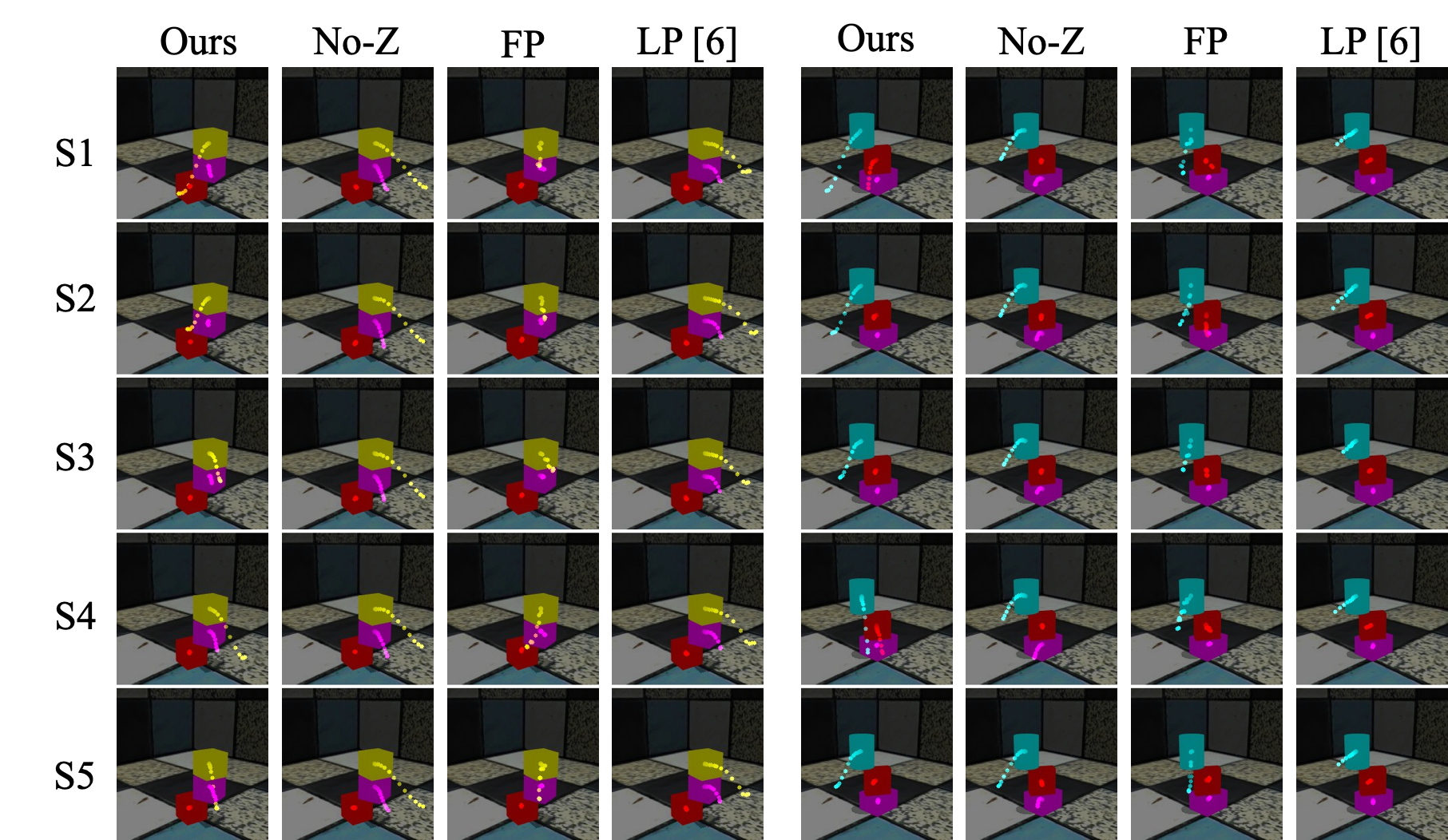}
    \caption{Visualizing five randomly sampled predictions by our method and other baselines. The predicted centers of entities over time overlay on top of the initial frame.}
    \label{fig:enc_vis}
    \vspace{-3mm}
\end{figure}

%% file: gym_vis.tex
\begin{figure*}[th]
    \centering
        \includegraphics[width=0.49\textwidth]{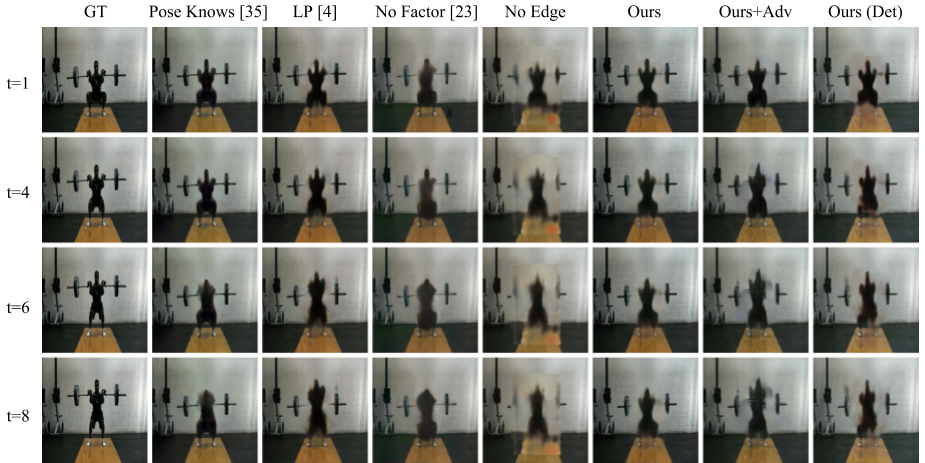}
        \includegraphics[width=0.49\textwidth]{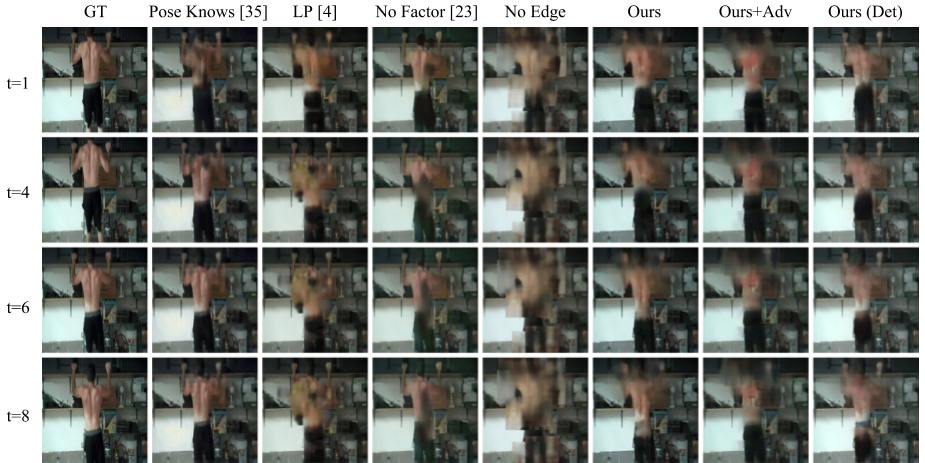}
    \caption{Video prediction results with best LPIPS latent using our approach compared to baselines. The last column visualizes results when the entity (joints) locations are replaced by the detection  in both training and testing. Videos are in supplementary.}
    \label{fig:gym_vis}
    \vspace{-3mm}
\end{figure*}

%% file: gym_sample.tex
\begin{figure}
    \centering
        \includegraphics[width=.9\linewidth]{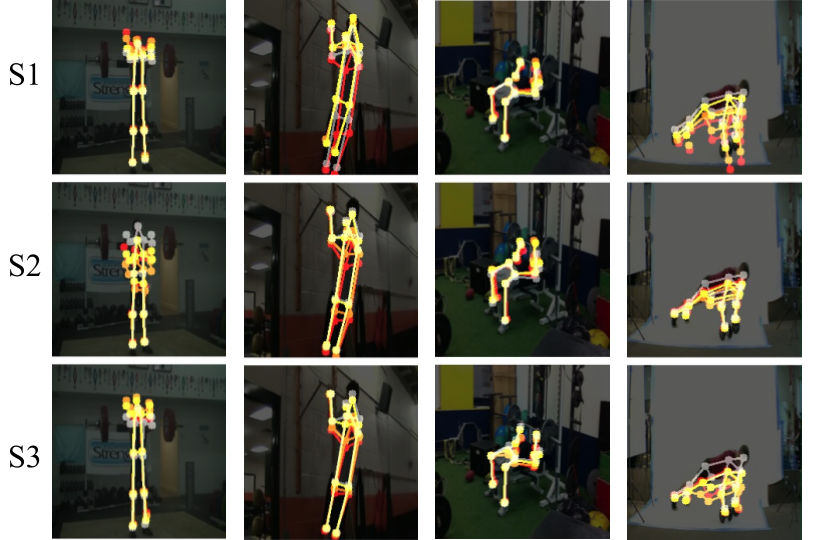}
    \caption{Visualizing joint positions in three randomly sampled predictions by our method. The initial skeletons are plotted as white. Skeletons at time 0.25s, 0.5s, and 1s are plotted as yellow, orange, and red, respectively.}\label{fig:gym_sample}
    \vspace{-1mm}
\end{figure}

%% file: gym_quan.tex
\begin{figure}[t!]
    \centering
        \includegraphics[width=\linewidth]{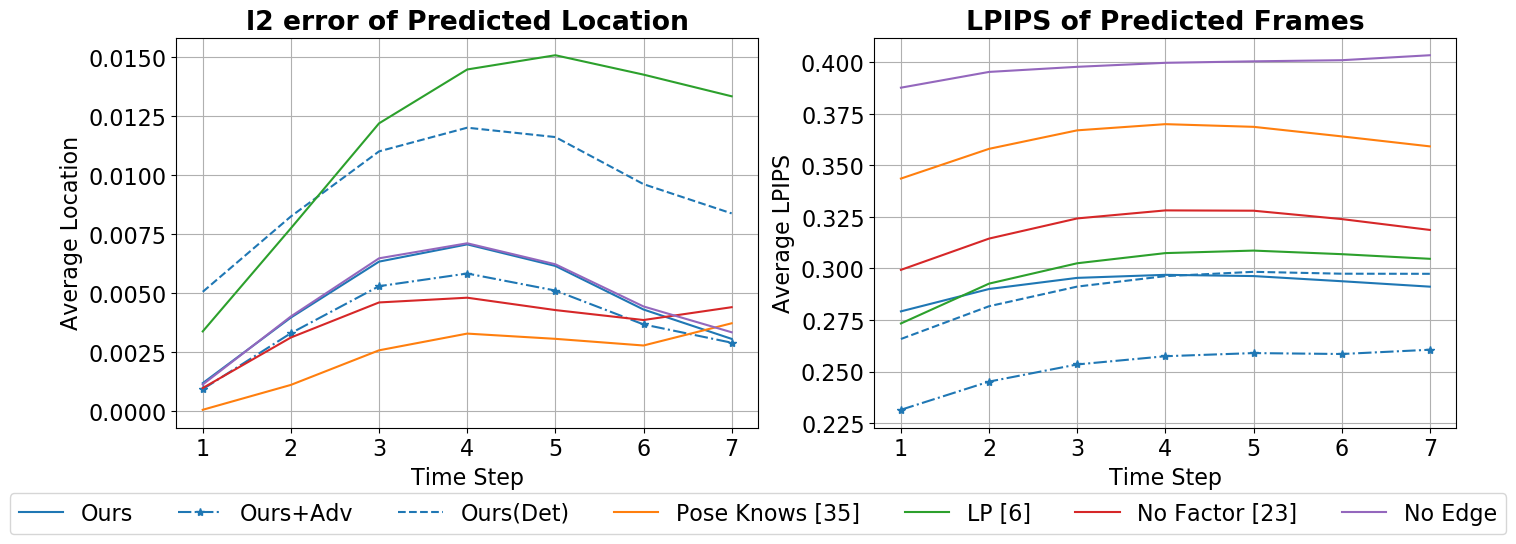}
    \caption{Error for location (Left) and frame (Right) prediction  using our model and baseline methods. For each sequence, the best score of 100 samples is drawn.}\label{fig:gym_quan}
    \vspace{-2mm}
\end{figure}

%% file: discussion.tex
In this work we proposed a method that leverages compositionality across entities for video prediction. However, the task of video prediction in a general setting is far from being solved, and many challenges still remain. In particular, we rely on supervision of the entity locations, either from human or automatic annotations. It would be interesting to relax this requirement and allow the entities to emerge as pursued in some recent works~\cite{chang2016compositional,hsieh2018learning}, although in simpler settings. Additionally, GAN-based auxiliary losses have been shown to improve image synthesis quality, and these could be explored in conjunction with our model. Lastly, developing metrics to evaluate the diversity and accuracy of predictions is also challenging due to the multi-modal nature of the task, and we hope some future efforts will also focus on this aspect.


%% file: appendix.tex
\section{Architecture Details}
\paragraph{Entity Predictor.}
Our predictor leverages the graph neural network family, whose learning process  can be abstracted to iterative message passing and message aggregation. In each round of message passing, each node  (edge) is a parameterized function of their neighboring node and edges, which updates their parameters by back propagation. We introduce the predictor architecture by instantiating the message passing and aggregation operation as following:

For the $l$-th layer of message passing, it consists of two operations: 
\begin{align*}
v\to e :& e^{(l)}_{i,j} = f^{(l)}_{v\to e}[v^{(l)}_i \oplus v^{(l)}_{j}] \\
e\to v:& v_{i}^{(l+1)} = f^{(l)}_{e\to v} [\text POOL[e_{i, j}^{(l)} | (i,j) 
\end{align*}
We first perform node-to-edge passing $f^{(l)}_{v\to e}$ where edge embeddings are implicitly learned. Then we perform edge-to-node $f^{(l)}_{v\to e}$ operation given the updated edge embeddings. The message passing block can be stacked to arbitrary layers to perform multiple rounds of message passing between edge and node. 
In our experiment, we stack four blocks of the above module. For each block, $f^{(l)}_{v\to e}, f^{(l)}_{e\to v}$ are both implemented as a single fully connected layer. The aggregation  operator is implemented as a average pooling. 
Note that connection expressed in the edge set can be either from explicitly specified graph, or a fully connected graph when the relationship is not explicitly observed.

\vspace{-1em}
\paragraph{Frame Decoder.}
We use the backbone of Cascaded Refinement Networks. Given feature in shape of ($N, D, h_0, w_0$) either from entity predictor or background feature, the frame decoder upsamples it at the end of every unit. Each unit comprises of $Conv \to Batch \to LeakyRelu$. When the entity features are warped to image coordinates, the spatial transformation is implemented as a forward transformation to sharpen entities.

\vspace{-1em}
\paragraph{Latent Encoder.}
At training, the encoder takes in the concatenated features of two frames and  apply a one layer neural network to obtain mean and variance of $u$, where we resample with reparameterization trick at training time. The resampled $u'$ is fed into a one-layer LTSM as cell unit to generates a sequence of $z^t$.

\vspace{-1em}
\paragraph{Training Details.}
We optimize the total loss with Adam optimizer in learning rate  $1e-4$. $\lambda_1= 100, \lambda_2 = 1e-3$. The dimensionality of latent is 8, i.e. $|u| = |z^t| = 8$. Location feature is represented as the center of entities $|b|=2$, appearance feature $|a| = 32$. The region of each entity is set to a large enough fixed width and height to cover the entity, $d=70$ in all of our experiment.  All generated frame are in resolution of $224\times 224$.

\section{Dataset}

In Shapestacks, the `entities' correspond to distinct objects, among which the graph used for interaction is fully connected since no explicit relationships are observed. 
The videos are generated by simulating the given initial configurations in in mujoco~\cite{todorov2012mujoco} for 16 steps. While the setting is deterministic under perfect state information (precise 3D position and pose, mass, friction, etc), the prediction task is ambiguous given an image input. The subset is split to 1320 clips for training, 281 clips for validation, and 296 clips for testing. When we evaluate the generalization ability, the test set further includes 221 (136 / 93) clips of videos comprised of 4 (5 / 6) blocks.

In Penn Action, `entities' correspond to joints of human body and the graph is built based on prior knowledge of skeletons. If some joint is missing in the video, we instead link the edge to its parent if possible.  We train our model to generate video sequences of 1 second at 8 fps given an initial frame. The categories we used are bench press, clean and jerk, jumping jacks, pull up, push up, sit up, and squat. To reduce overfitting, we augment data on the fly, including randomly selecting the starting frame for each clip, random spatial cropping, etc.

\section{Baseline Model}
The No-Factor model does not predict a per-entity appearance but simply outputs a global feature that is decoded to foreground appearance and mask.  To ensure the use of the same supervision as box locations, the No-Factor model also takes as input (and outputs) the per-entity bounding boxes. Thus, the foreground is represented as the extracted feature of the entire frame concatenated by all locations and they are directly predicted together with fully connected layers. To decode them to pixels, an additional binary mask is applied. However, no mechanism in No-Factor baseline guarantees the generated pixels of entities respect the predicted locations. 

In the No-Edge baseline, we remove all but self-link edges between nodes so that all the nodes are predicted independently.

Pose-Knows~\cite{walker2017pose} consists of two models: a Pose-VAE that takes input as the initial frame together with joint location and outputs joint location in the future, a Pose-GAN with skip layers that takes input as the initial frame together with rendered predicted poses and generate frames. The original work uses 3D convolutional ~\cite{tran2015learning} network to generate low resolution videos ($80\times 64$). However, with progress in GAN techniques in recent years ~\cite{wang2018high}, we find that 2D convolution with frame-wise adversarial loss improves performance when generating high resolution videos ($224\times224$) in terms of both qualitative and quantitative evaluation.   

\section{Standard Deviation}
\begin{figure} [t]
    \centering
    \includegraphics[width=\linewidth]{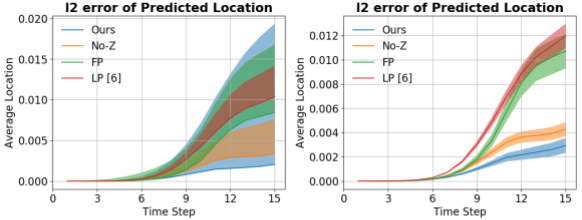}
    \caption{Left: For all 100 samples, $\sigma$ (shade) and  best samples (line at lower boundary of shade) are plotted. Right: For best 5 samples of 100, mean and $\sigma$ are plotted. }
    \label{fig:var}
\end{figure}

Prediction task is multi-modal, a model that correctly handles uncertainty will predict diverse future states (and therefore should) in the error across samples (as $\sigma=0$ implies mean prediction). while generate enough good samples.
We plot the $\sigma$ in location error (Shapestacks) over 100 samples in Figure \ref{fig:var} (Left).  We also report the $\sigma$ across top 5 samples in Figure \ref{fig:var} (right) 